\documentclass[manuscript,screen,natbib=false]{acmart}

\usepackage{cite}
\usepackage{amsmath,amssymb,amsfonts}
\usepackage{algorithmic}
\usepackage[ruled]{algorithm2e}
\usepackage{graphicx}
\usepackage{textcomp}
\usepackage{setspace} 
\usepackage{xcolor, soul}
\usepackage{dashrule}
\usepackage{lineno}

\AtBeginDocument{%
  \providecommand\BibTeX{{%
    \normalfont B\kern-0.5em{\scshape i\kern-0.25em b}\kern-0.8em\TeX}}}

\setcopyright{acmcopyright}
\copyrightyear{2025}
\acmYear{2025}
\acmDOI{XXXXXXX.XXXXXXX}

\begin{document}

\title[FedSAF: A Federated Learning Framework]{FedSAF: A Federated Learning Framework for Enhanced Gastric Cancer Detection and Privacy Preservation}

\author{YUXIN MIAO}
\author{XINYUAN YANG}
\author{HONGDA FAN}
\author{YICHUN LI}
\author{YISHU HONG }
\author{XIECHEN GUO}
\affiliation{%
  \institution{The University of Sydney}
  \country{Australia}
}
\author{Ali Braytee}
\author{Weidong Huang}
\affiliation{%
  \institution{University of Technology Sydney}
  \country{Australia}
}

\author{Ali Anaissi}
\affiliation{%
  \institution{The University of Sydney}
  \country{Australia}
}
\affiliation{%
  \institution{University of Technology Sydney}
  \country{Australia}
}

\renewcommand{\shortauthors}{}

\begin{abstract}

Gastric cancer is one of the most commonly diagnosed cancers and has a high mortality rate. Due to limited medical resources, developing machine learning models for gastric cancer recognition provides an efficient solution for medical institutions. However, such models typically require large sample sizes for training and testing, which can challenge patient privacy. Federated learning offers an effective alternative by enabling model training across multiple institutions without sharing sensitive patient data. This paper addresses the limited sample size of publicly available gastric cancer data with a modified data processing method. Among existing pre-trained models, select the most suitable local computer vision model for gastric cancer classification. Then the paper introduces FedSAF, a novel federated learning algorithm designed to improve the performance of existing methods, particularly in non-independent and identically distributed (non-IID) data scenarios. FedSAF incorporates attention-based message passing and the Fisher Information Matrix to enhance model accuracy, while a model splitting function reduces computation and transmission costs. Hyperparameter tuning and ablation studies demonstrate the effectiveness of this new algorithm, showing improvements in test accuracy on gastric cancer datasets, with FedSAF outperforming existing federated learning methods like FedAMP, FedAvg, and FedProx. The framework's robustness and generalization ability were further validated across additional datasets (SEED, BOT, FashionMNIST, and CIFAR-10), achieving high performance in diverse environments. Overall, the improved data processing method and the FedSAF algorithm provide a data-centric solution for gastric cancer research and contribute significantly to the development of federated learning. The well-trained FedSAF can be applied in practice to protect patient privacy while offering a reliable framework for collaborative medical institutions. This innovative algorithm also introduces a new perspective for data types requiring private information sharing. Future federated learning research can build upon the methodologies proposed in this paper.

\end{abstract}


\maketitle

\section{Introduction}
\label{sec:introduction}

Gastric cancer is one of the most prevalent and deadly cancers worldwide, contributing significantly to global mortality rates. Its recurrence and mortality rates remain alarmingly high year-round, posing a substantial threat to public health. According to the World Health Organization (WHO), millions of people die from cancer annually, with gastric cancer being a leading contributor to these deaths \cite{sexton2020gastric}. In recent years, deep neural networks (DNNs) have shown promising results in detecting anomalies in medical imaging, becoming widely used for early diagnosis and detection of various cancer types \cite{mireshghallah2020privacy}. However, due to the highly sensitive nature of medical data and stringent privacy protection regulations, hospitals are often unable to share patient data for centralized training \cite{pfitzner2021federated}. This limitation hampers the effectiveness of model training and recognition accuracy, potentially delaying timely interventions for patients.

To address this issue, this paper proposes an efficient gastric cancer recognition framework aimed at improving diagnostic accuracy while protecting data privacy and fostering collaboration between medical institutions. Federated Learning (FL), an emerging distributed machine learning technique, offers a promising solution to these challenges. FL enables each medical institution to train the model locally, sharing only model parameters rather than the original data. After aggregating the parameters through a weighted process, the server generates an updated global model and distributes it to each client for the next round of training. Through multiple rounds of iterations, federated learning integrates the learning capabilities of multi-source data while safeguarding patient privacy, thereby significantly enhancing the generalization ability and predictive performance of the model \cite{li2020federated}.

Despite its potential, gastric cancer diagnosis models built on federated learning face several challenges in real-world applications. First, medical data is often non-independent and identically distributed (non-iid), meaning that the data varies due to factors like differences in medical equipment across institutions or patient population heterogeneity. This variability can negatively impact the model's generalization ability and diagnostic precision \cite{ma2022nonIID}. Furthermore, traditional averaging methods for aggregating model parameters may not fully leverage the critical data from institutions with more important or specialized datasets. This paper addresses these issues by introducing a client importance evaluation mechanism that assigns higher weights to more valuable data, improving the global model’s accuracy. To further enhance the model, a model splitting mechanism is proposed to ensure the personalization of local models while retaining global consistency. This helps preserve the unique characteristics of individual institutions’ data, fostering better model performance. Moreover, federated learning incurs high communication costs as data is exchanged between institutions. This paper also seeks to address this issue by optimizing communication efficiency, and ensuring the privacy of data transmission while reducing the computational overhead \cite{mcmahan2017communication}.

The ultimate goal of this paper is to develop a robust and flexible FL framework that improves the accuracy and efficiency of gastric cancer detection. By optimizing data utilization and enhancing privacy-preserving collaboration across medical institutions, this framework seeks to provide reliable diagnostic tools that could significantly lower the global mortality rate of gastric cancer through earlier and more precise detection.

In this paper, we propose the following main contributions:

\begin{itemize}
    \item \textbf{Federated Learning Framework}: Proposed a federated learning model to improve gastric cancer detection while ensuring data privacy.
    \item \textbf{Model Splitting Mechanism}: Developed a model splitting mechanism to retain personalized local model parameters while maintaining global consistency.
    \item \textbf{Client Importance Evaluation}: Introduced a client importance evaluation mechanism to assign higher weights to more valuable data, optimizing model performance.
    \item \textbf{Data Heterogeneity Handling}: Addressed the challenge of non-iid data across medical institutions, improving model generalization and diagnostic accuracy.
\end{itemize}

\section{Related work}

Machine learning has become a transformative tool in medical science, playing a vital role in disease warning, prediction, and supporting medical diagnoses \cite{Chen-Zhang-Che-Huang-Han-Yuan-2021}. Among the many challenges in this domain, cancer remains a leading cause of mortality, posing a significant threat to human health. Leveraging machine learning to improve cancer cell identification has thus become a critical area of research. However, the practical implementation of such techniques faces several challenges, particularly due to the sensitive nature of medical data and the strict privacy regulations that govern its usage \cite{agcaoglu2020dynamic, inkster2018china}. These regulations, coupled with uneven distribution of medical resources across regions, result in highly heterogeneous cancer datasets characterized by significant variations in size, quality, and distribution \cite{guo2023breast}. Consequently, traditional machine learning approaches struggle to achieve satisfactory performance when applied to limited, private datasets.



The traditional goal of FL is to train a generic global model suitable for all clients. While this approach is effective in scenarios with consistent data distribution, it falters in handling heterogeneous datasets. Personalized Federated Learning (pFL) has emerged as a solution to this problem, aiming to balance global collaboration with local personalization. By training a local model for each client, pFL ensures that the models are better suited to the unique characteristics of the local data. Methods such as pFedMe and Ditto exemplify this approach. pFedMe employs the Moreau envelope to smooth the objective function and optimize both global and personalized models, leading to faster convergence and superior performance on non-IID data \cite{t2020personalized}. Ditto introduces a regularization term to balance global and personalized objectives, offering flexibility in controlling the degree of personalization based on hyperparameters \cite{li2021ditto}. However, these methods often come with increased computational complexity and require careful tuning to address severe data heterogeneity effectively.

Another promising approach within pFL is personalized aggregation. This method adjusts aggregation weights dynamically based on the unique characteristics of each client. FedAMP, for example, leverages an attention message passing mechanism to adaptively adjust parameter sharing among clients with similar data distributions \cite{huang2021personalized}. Similarly, FedApple and FedAS introduce advanced techniques to tackle cross-domain adaptation and data inconsistency, though these methods often demand higher computational resources \cite{luo2022adapt, yang2024fedas}. While personalized aggregation enhances collaboration and personalization, it may compromise efficiency in resource-constrained settings.

Model-splitting-based pFL offers another innovative direction by dividing models into shared and personalized layers. FedPer, for instance, ensures global collaboration through shared layers while preserving local data characteristics via personalized layers \cite{arivazhagan2019federated}. This approach effectively balances global and local objectives while reducing communication costs. However, fixed layer selection can limit flexibility, potentially hindering personalization for diverse client needs.

The advancements in FL and pFL have significant implications for cancer image recognition tasks. These methods not only address the challenge of insufficient sample sizes caused by resource imbalances but also ensure robust privacy protection. While traditional FL excels at creating stable global models, pFL strikes a balance between personalization and global collaboration, offering superior performance under non-IID conditions. However, challenges related to communication costs, computational efficiency, and personalization remain areas of active research.

Building on insights from the existing literature, this paper proposes an innovative FL framework to address the challenges of heterogeneity and privacy in medical data. 

\section{Proposed Method/Framework}

This section introduces a novel algorithm, Federated Split-Aware Framework (FedSAF), designed to address the challenges of data heterogeneity, limited communication efficiency, and the need for personalization in federated learning systems. FedSAF incorporates three key components: Model Splitting, Attentive Message Passing, and the Fisher Information Matrix (FIM). Together, these components form a cohesive framework that enhances model performance in non-IID (non-independent and identically distributed) data environments while preserving computational efficiency and data privacy.

\subsection{Model Splitting}



The model-splitting technique employed in FedSAF 
plays a pivotal role in significantly reducing communication costs. This technique divides the model into two distinct parts: shared global layers and personalized local layers. 
\begin{itemize}
    \item Shared Global Layers: These layers are designed to capture universal features that are consistent across all clients. 
    They facilitate efficient collaboration by transmitting global knowledge among clients, ensuring effective feature sharing and reducing redundancy in the representation of common patterns.
   
    \item Personalized Local Layers: These layers are independently trained on each client’s local data, enabling the model to adapt to specific data characteristics. This personalized component ensures that the framework addresses the heterogeneity of data distributions across clients.
\end{itemize}

The detailed formulation of the model-splitting operation is presented in Equation~\ref{eq:base_calculation_appendix}.

\begin{equation}
    \rho_i = \tau_i - \eta_i 
    \label{eq:base_calculation_appendix}
\end{equation}
where:
\begin{itemize}
    \item $\rho_i$ represents the shared layers of client $i$,
    \item $\tau_i$ denotes the complete model parameters of client $i$,
    \item $\eta_i$ denotes the personalized head layers of client $i$.
\end{itemize}

In this approach, only the parameters of the shallow neural network layers (represented by $\rho_i$) are shared with the server, while the deeper network layers (represented by $\eta_i$) remain personalized and fixed on the client side. By reducing the total model parameters $\tau_i$ to only the shared layers $\rho_i$ and freezing the personalized head layers $\eta_i$ locally, the amount of parameters exchanged between the client and server is significantly minimized.

By leveraging the model-splitting technique, FedSAF achieves a balance between global collaboration and local personalization. This decoupling of shared and local layers not only addresses client data variability but also significantly reduces communication overhead by limiting the transmission to updates from the global layers. Consequently, this enhances the overall efficiency of the framework.

\subsubsection{Hyperparameter Tuning in Model Splitting}

An essential aspect of the model-splitting strategy is the tuning of the  hyperparameter \( nhead \), which determines the number of layers shared or fixed locally. This parameter controls the depth of the personalized layers (\( \eta_i \)) versus the shared layers (\( \rho_i \)) for each client. For instance, using the MobileNetV3 Small architecture as an example, the structure is outlined in Table~\ref{tab:mobilenetv3_structure}. 

\begin{figure}[!t]
    \centering
    \begin{minipage}{0.3\textwidth}
        \centering
        \includegraphics[width=\textwidth]{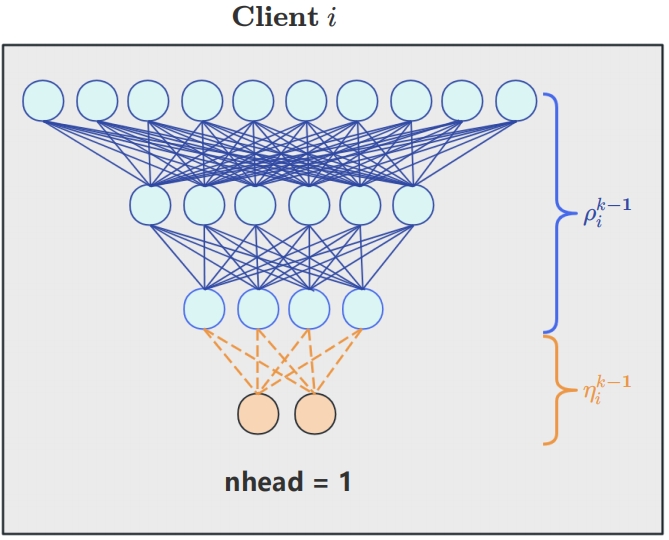}
        \Description{Subfigure (a) shows the results of the model when nhead is set to 1.}
        \caption*{(a) nhead:1}
    \end{minipage}
    \begin{minipage}{0.3\textwidth}
        \centering
        \includegraphics[width=\textwidth]{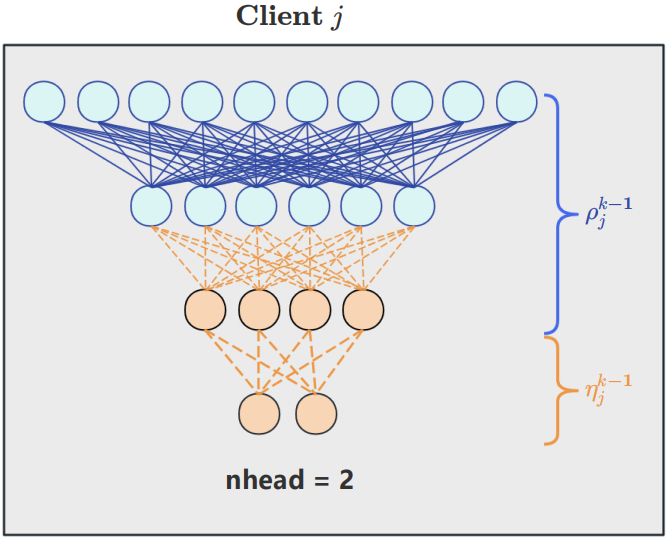}
        \Description{Subfigure (b) shows the results of the model when nhead is set to 2.}
        \caption*{(b) nhead:2}
    \end{minipage}
    \begin{minipage}{0.3\textwidth}
        \centering
        \includegraphics[width=\textwidth]{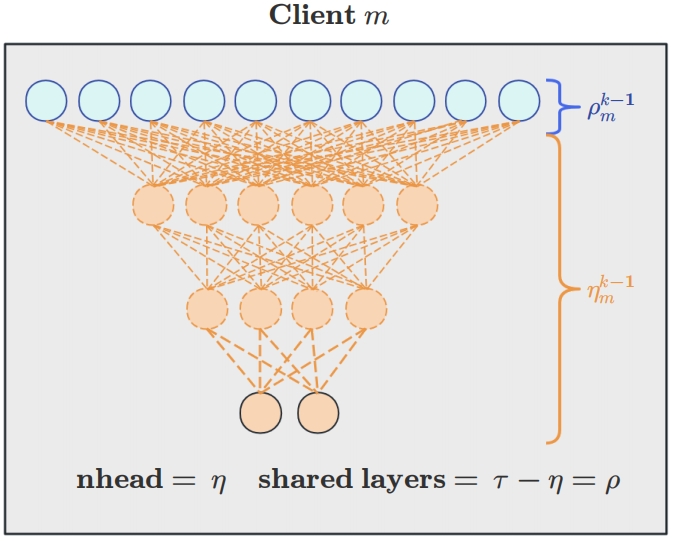}
        \Description{Subfigure (c) shows the results of the model when nhead is set to $\rho$.}
        \caption*{(c) nhead:$\rho$}
    \end{minipage}
    \Description{This figure shows the results of hyperparameter tuning for different values of nhead (1, 2, and $\rho$).}
    \caption{Hyperparameter Tuning for Different Values of nhead.}
    \label{fig:nhead_combined}
\end{figure}

As illustrated in Figure~\ref{fig:nhead_combined}, varying \( nhead \) determines which layers are fixed locally versus shared globally.  
\begin{itemize}
    \item When \( nhead = 1 \), only the fully connected (fc) layer is fixed, and the remaining layers are shared.
    \item When \( nhead = 2 \), the fc layer and the \(ConvBNReLU\) layer at position 14 are fixed,while the remaining layers are shared.
    \item For \( nhead = \eta \), the first \(\eta\) layers are fixed locally, while the remaining \(\rho\) layers are shared.
\end{itemize}

The rationale for layer splitting is  rooted in the observation that lower layers in neural networks typically learn basic features such as edges and corners, while higher layers capture more complex and abstract representations \cite{yosinski2015understanding}. Sharing parameters from the lower layers enables the global model to generalize effectively across clients, as these features are common across different datasets. Conversely, higher layers often capture domain-specific features, making them more suitable for personalization at the client level. The training strategy alternates between updating the head and base layers, as follows:
\begin{enumerate}
    \item \textbf{Step 1:} Freeze base layers and update head layers to capture specific patterns in local data.
    \item \textbf{Step 2:} Freeze head layers and update base layers to ensure robust global feature learning.
\end{enumerate}

This alternating approach ensures efficient local model training while reducing computational and communication costs. Additionally, only the base layer parameters are transmitted to the server, further optimizing the overall training process.

By fine-tuning \( nhead \), FedSAF achieves a customizable balance between generalization and personalization, effectively catering to diverse data distributions among clients.

\begin{table}[!t]
    \centering
    \caption{Example: Structure of MobileNetV3 small}
    \begin{tabular}{ccc}
        \toprule
        \textbf{Layer ID} & \textbf{Main Layer} & \textbf{Details} \\
        \midrule
        0-2 & features & ConvBNReLU, BatchNorm2d, ReLU6 \\
        3-13 & features & InvertedResidual (layers with ConvBNReLU, BatchNorm2d, ReLU6) \\
        14 & features & ConvBNReLU \\
           & dropout & Dropout(p=0.2, inplace=False) \\
        15 & fc & Linear(in\_features=1280, out\_features=2, bias=True) \\
        \bottomrule
    \end{tabular}
    \label{tab:mobilenetv3_structure}
\end{table}

\subsection{Attentive Message Passing (AMP)}
To further improve personalization, FedSAF employs an Attentive Message Passing (AMP) mechanism during the aggregation stage. The AMP component dynamically adjusts the contribution of each client to the global model based on the similarity of their data distributions. Key components of the AMP mechanism are:

\begin{itemize}
    \item \textbf{Parameter Similarity Weighting:} FedSAF uses the distance between model parameters to measure the similarity of local data distributions. Clients with similar data distributions are assigned higher weights during aggregation, promoting collaboration among clients with comparable data characteristics.
    \item \textbf{Adaptive Aggregation:} The attentive mechanism ensures that the global model prioritizes relevant updates, improving the quality of the aggregated model. This approach not only enhances model convergence but also facilitates better personalization for each client.
\end{itemize}

AMP enables FedSAF to balance the trade-off between generalization and personalization, ensuring robust performance across a wide range of heterogeneous environments. 


The intermediate model \( z_i^k \) (defined in Equation~\ref{AMP server appendi}) of each client \( i \) is a weighted average of its own model and those of other clients.

\begin{equation}
    z_i^k = \xi_{i,i} \tau_i^{k-1} + \sum_{j \neq i}^{m} \xi_{i,j} \tau_j^{k-1}
    \label{AMP server appendi}
\end{equation}

Where:
\begin{itemize}
    \item \( z_i^k \) is the intermediate model for client \( i \) at communication round \( k \),
    \item \( \xi_{i,j} \) represents the weight assigned to the model parameters from client \( j \) when updating client \( i \)'s model,
    \item \( \tau_i^{k-1} \) denotes the model parameters of client \( i \) from the previous round \( k-1 \).
\end{itemize}

\subsubsection{AMP Used in the Server and Hyperparameter Tuning}

During server aggregation, the degree of collaboration between clients is depended on the similarity of their model weights. There are three distance formulas used to calculate the similarity of model weights: cosine distance, Euclidean distance and Manhattan distance. The distance formula is set as a hyperparameter, abbreviated as \textbf{dm}.

The distance formula affects the weight coefficients of other clients (e.g., client \( j \)) used for collaboration when aggregating the global model on the server. For Euclidean Distance, the formulas in Equations~\ref{eq:xi_Euclidean_total} and ~\ref{eq:xi_Euclidean_base} refer to the weight coefficients without and with model segmentation, respectively. The  Euclidean distance  between Client$_{i}$ and Client$_{j}$  is expressed as:

If not splitting model:
\begin{equation}
    \xi_{i,j} = \alpha_k A'\left( \left\| \mathbf{\tau}_i^{k-1} - \mathbf{\tau}_j^{k-1} \right\|^2 \right), \quad (i \neq j)
    \label{eq:xi_Euclidean_total}
\end{equation}

If splitting model:

\begin{equation}
    \xi_{i,j} = \alpha_k A'\left( \left\| \mathbf{\rho}_i^{k-1} - \mathbf{\rho}_j^{k-1} \right\|^2 \right), \quad (i \neq j)
    \label{eq:xi_Euclidean_base}
\end{equation}

Where:
\begin{itemize}
    \item $\xi_{i,j}$ is the similarity measure between client $i$ and client $j$, using model parameters from the previous round $k-1$.
    \item $\alpha_k$ is a scaling factor for round $k$, which adjusts the influence of the distance measure.
    \item $A'(\cdot)$ is a function transforming the distance measure, applied to the squared Euclidean distance.
    \item $\mathbf{\tau}_i^{k-1}$ and $\mathbf{\rho}_i^{k-1}$ represent the parameter vectors from client $i$ for the non-splitting and splitting models, respectively, from the previous round $k-1$.
    \item $\mathbf{\tau}_j^{k-1}$ and $\mathbf{\rho}_j^{k-1}$ are the parameter vectors from client $j$ affecting the update of client $i$, corresponding to the non-splitting and splitting models respectively.
    \item $\left\| \mathbf{\tau}_i^{k-1} - \mathbf{\tau}_j^{k-1} \right\|^2$ and $\left\| \mathbf{\rho}_i^{k-1} - \mathbf{\rho}_j^{k-1} \right\|^2$ are the squared Euclidean distances between the parameter vectors of clients $i$ and $j$ for the non-splitting and splitting models, respectively.
\end{itemize}

Similarly, for Manhattan Distance, the formulas in Equations~\ref{eq:xi_manhattan_total} and ~\ref{eq:xi_manhattan_base} apply to the weight coefficients with and without model segmentation, respectively.  The  Manhattan distance  between Client$_{i}$ and Client$_{j}$ is expressed as:

If not splitting model:
\begin{equation}
    \xi_{i,j} = \alpha_k A'\left( \left| \mathbf{\tau}_i^{k-1} - \mathbf{\tau}_j^{k-1}\right| \right), \quad (i \neq j)
    \label{eq:xi_manhattan_total}
\end{equation}

If splitting model:
\begin{equation}
    \xi_{i,j} = \alpha_k A'\left( \left| \mathbf{\rho}_i^{k-1} - \mathbf{\rho}_j^{k-1}\right| \right), \quad (i \neq j)
    \label{eq:xi_manhattan_base}
\end{equation}


For Cosine Distance, the formulas in quations~\ref{eq:xi_cosine_total} and ~\ref{eq:xi_cosine_base} describe the weight coefficients for both cases. The cosine similarity to measure the alignment of the model parameter vectors between Client$_{i}$ and Client$_{j}$ is expressed as:

If not splitting model:
\begin{equation}
    \xi_{i,j} = \alpha_k A'\left(\frac{\mathbf{\tau}_i^{k-1} \cdot \mathbf{\tau}_j^{k-1}}{\|\mathbf{\tau}_i^{k-1}\| \|\mathbf{\tau}_j^{k-1}\| + 1e-8}\right), \quad (i \neq j)
    \label{eq:xi_cosine_total}
\end{equation}

If splitting model:
\begin{equation}
    \xi_{i,j} = \alpha_k A'\left(\frac{\mathbf{\rho}_i^{k-1} \cdot \mathbf{\rho}_j^{k-1}}{\|\mathbf{\rho}_i^{k-1}\| \|\mathbf{\rho}_j^{k-1}\| + 1e-8}\right), \quad (i \neq j)
    \label{eq:xi_cosine_base}
\end{equation}

\subsubsection{AMP Used in the Client}

On the client side, AMP is used as part of the regularization for $\mathcal{G}(\tau)$, which is the global objective function to be minimized. $\mathcal{G}(\tau)$  is used to update model parameters \(\tau\). The equations of AMP used in the aggregation in the client are expressed by:

\begin{equation}
    \min_\tau \left\{ \mathcal{G}(\tau) := \sum_{i=1}^{m} F_i(\tau_i) + \lambda \sum_{i<j} A(\|\mathbf{\tau}_i - \mathbf{\tau}_j\|^2) \right\}
    \label{eq:objective_function}
\end{equation}

Where \( \mathcal{G}(\tau) \) consists of two components: the client \( i \)'s loss function and the regularization term. 

Equation~\eqref{eq:loss_function} represents the loss \( F_i(\tau) \) on client \( i \), which measures model performance by comparing the prediction \( f(\tau, x_n) \) of model \( f \) using parameters \( \tau \) on input \( x_n \) with the label \( Y_n \).

Additionally, \( A(\cdot) \) is a function applied to the squared Euclidean distance between model parameters of different clients. The attentive function \( A(\cdot) \) employs a negative exponential function, as shown in Equation~\ref{eq:attention_function}.

\begin{equation}
    A(\|\mathbf{\tau}_i - \mathbf{\tau}_j\|^2) = 1 - e^{-\|\mathbf{\tau}_i - \mathbf{\tau}_j\|^2 / \sigma},
    \label{eq:attention_function}
\end{equation}

Where:
\begin{itemize}
    \item When $\|\mathbf{\tau}_i - \mathbf{\tau}_j\|^2$ is small, the term $e^{-\|\mathbf{\tau}_i - \mathbf{\tau}_j\|^2 / \sigma}$ approaches 1, leading $A(\|\mathbf{\tau}_i - \mathbf{\tau}_j\|^2)$ to approach 0.
    \item When $\|\mathbf{\tau}_i - \mathbf{\tau}_j\|^2$ increases, the term $e^{-\|\mathbf{\tau}_i - \mathbf{\tau}_j\|^2 / \sigma}$ gradually approaches 0, making $A(\|\mathbf{\tau}_i - \mathbf{\tau}_j\|^2)$ gradually increase towards 1.
\end{itemize}
 The regularization term $\lambda$ on the left acts as a penalty component, and the loss function \( F_i(\tau) \) is defined as:

\begin{equation}
    F_i(\tau) = \frac{1}{N} \sum_{n=1}^{N} \ell\left(f(\tau, x_n), Y_n\right)
    \label{eq:loss_function}
\end{equation}

Where:
\begin{itemize}
    \item $F_i(\tau)$ is the loss function for client $i$.
    \item $\ell$ is the loss function, which is the cross-entropy loss in this case.
    \item $f(\tau, x_n)$ is the prediction result of the model $f$ using parameters $\tau$ for the input $x_n$.
    \item $Y_n$ is the label for the input $x_n$.
\end{itemize}

As seen in Equation~\eqref{eq:objective_function}, the regularization term serves to reduce model parameter differences between clients. When the differences between two clients' model parameters are large, the regularization focuses on minimizing the variation. As \( \mathcal{G}(\tau) \) evolves, the optimization task is to minimize the overall objective function. The optimization process aims to smooth the model differences across clients, reducing variations among them. The loss function \( F_i(\tau_i) \) focuses on the local model of each client, aligning each client’s model with the global model while accounting for local data. The regularization term ensures consistency by reducing the model parameter differences across clients. A larger \( \lambda \) indicates a stronger penalty effect, improving alignment across clients' models, but if \( \lambda \) is too large, it may suppress the unique local characteristics of each client's model.

 \subsection{Fisher Information Matrix (FIM)}
The Fisher Information Matrix (FIM) is integrated into FedSAF to evaluate the global contribution of client parameters and enhance the relevance of shared information. FIM measures the importance of each parameter in the global model by assessing its sensitivity to the likelihood of observed data.

\begin{itemize}
    \item \textbf{Global Contribution Measurement:} By analyzing the FIM, FedSAF identifies the parameters that are most critical for global model optimization, ensuring that significant contributions are prioritized during aggregation.
    
    \item \textbf{Regularization for Stability:} The FIM also serves as a regularization tool, guiding the model toward stable convergence by minimizing the impact of irrelevant or noisy updates.
\end{itemize}

Incorporating FIM allows FedSAF to achieve a higher degree of collaboration between clients without compromising the quality of personalized models.




In summary, inadequately trained stragglers hinder model aggregation and create inconsistencies with other normally trained clients. Therefore, it is crucial to mitigate the negative impact of stragglers on the aggregation process.

The Fisher Information Matrix (FIM) is an efficient method to address this issue. Statistically, FIM is defined as the expected value of the second partial derivative of the Negative Log-Likelihood (NLL) with respect to the model parameters, measuring the sensitivity of the prediction outcomes to variations in those parameters. A higher FIM value indicates that the corresponding parameter matrix plays a more significant role in contributing to the global model. Therefore, a straggler with a low FIM value will receive a lower aggregation weight, thereby minimizing its negative impact on the overall model.

\subsubsection{FIM Used in the Client}

The Fisher Information Matrix (FIM) is computed at the client level, where a low FIM value indicates that a client is an under-trained straggler.

To address the challenges of federated learning, particularly the highly imbalanced (non-iid) data distributions, Negative Log-Likelihood (NLL) combined with the softmax function is used instead of the sigmoid function. This combination improves probability distinctions between classes and ensures FIM, derived from NLL, accurately reflects the sensitivity of model parameters to prediction errors.

FIM faithfully reflects the steepness of the NLL gradient and quantifies the curvature elasticity of NLL concerning the model parameters \( \theta \). The formula for FIM is expressed as:

\begin{equation}
    \text{FIM}_i(\theta) = \frac{\partial^2 \text{NLL}_i}{\partial \theta \, \partial \theta^\top},
    \label{eq:fim_client}
\end{equation}
where:
\begin{itemize}
     \item \( \text{NLL}_i \) represents the Negative Log-Likelihood for client \( i \).
     \item \( \theta \) denotes the model parameters.
     \item \( \frac{\partial^2 \text{NLL}_i}{\partial \theta \, \partial \theta^\top} \) is the Hessian matrix of second-order derivatives of the NLL with respect to \( \theta \).
 \end{itemize}

To simplify the calculations, the trace of FIM is used in place of the full FIM. The trace focuses solely on the sum of the squared gradients of the NLL for each model parameter. In other words, the diagonal elements are prioritized. The formula for the trace of the Fisher Information Matrix (FIM) is expressed as follows:

\begin{equation}
    \text{trace}(\text{FIM}) = \sum_{i=1}^{n} \|\nabla_{\theta_i} \text{NLL}\|^2,
    \label{eq:fim_trace}
\end{equation}
where:
\begin{itemize}
    \item \( \text{trace}(\text{FIM}) \) denotes the trace of the Fisher Information Matrix.     \item \( n \) is the number of model parameters.
    \item \( \nabla_{\theta_i} \text{NLL} \) represents the gradient of the Negative Log-Likelihood with respect to the \( i \)-th parameter \( \theta_i \).
    \item \( \|\nabla_{\theta_i} \text{NLL}\|^2 \) is the squared norm of the gradient for the \( i \)-th parameter.
\end{itemize}

For FedSAF, on the client side, the trace of FIM is calculated using a mini-batch of private data. 
The trace of FIM is calculated inside each client and uploaded to the server as a list. Then, when the server aggregates the parameters, the weighted average of the trace of FIM is used as a coefficient to evaluate whether each client is sufficiently trained. 




\subsubsection{FIM Used in the Server}

During server aggregation, the FIM of each model can be used as a weighting coefficient, which helps reduce the participation of stragglers while enhancing the contributions of clients with richer informative content. The values of FIM serve as an important indicator for identifying untrained clients (stragglers). When a client \( i \) has a small trace of FIM, its weighted average coefficient (\(\gamma_i\)) will be lower than that of other clients. 

And the trace of FIM (tFIM), due to its computational efficiency, has been used as a practical substitute for FIM and is calculated on the client side. The equation for the weighted average coefficient of the tFIM function is as follows:

\begin{equation}
    \gamma_i = \frac{tFIM_i}{\sum_{n=1}^N tFIM_n}
    \label{eq:tfim_weight}
\end{equation}
where
\begin{equation}
    tFIM_i \leftarrow \text{trace of FIM} = \sum_{n=1}^{d_w} \text{diag}(FIM)_n
    \label{eq:tfim_trace}
\end{equation}

Here, \(\gamma_i\) represents the weight coefficient derived from the tFIM values, balancing the contribution of each client's model updates.

\subsection{Main Structure and Algorithm of FedSAF}

\subsubsection{Main Structure}

\begin{figure}[!t]
    \centering
    \includegraphics[width=\textwidth]{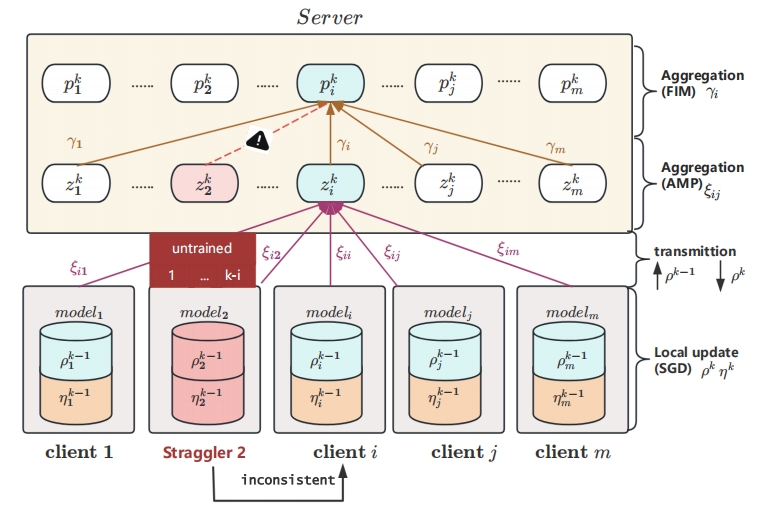}
    \Description{This is an image showing the structure of FedSAF, which illustrates the components and workflow of the algorithm.}
    \caption{The Structure of FedSAF}
    \label{fig:FedSAF}
\end{figure}

As shown in figure \ref{fig:FedSAF}, there are three core components of FedSAF, called \textit{Model Splitting}, \textit{AMP},\textit{ FIM}. In details, \textit{Model Splitting} occurs on the client side to reduce transmission costs. As shown in Figure~\ref{fig:FedSAF}, the client is divided into two parts. The blue cylinder represents the Shared Global Layers, while the orange cylinder represents the Personalized Local Layers. Only the Shared Global Layers are transmitted to the server for aggregation, while the Personalized Local Layers remain on the client for local updates. The hyperparameter \textit{the number of non-shared layers (nhead)} can be adjusted to control the number of Personalized Local Layers. For \textit{AMP}, it constitutes the first aggregation step on the server, designed to enhance collaboration among clients. After the first step, an intermediary variable $z_i^k$ is generated, which will be used for the next aggregation step. The hyperparameter \textit{distance methods (dm)} can be adjusted to change the degree of regularization of AMP. For \textit{ FIM}, it constitutes the second aggregation step on the server, aimed at reducing the negative impact of stragglers. After the second aggregation, an intermediary variable $p_i^k$ is produced. Notably, the FIM is calculated on the client side, where underperforming clients will receive lower FIM values. For example, \(Straggler_2\) in Figure \ref{fig:FedSAF}
represents an underperforming client. During previous training rounds (k-1), \(Straggler_2\) exhibited a significantly slower training speed compared to other clients, effectively remaining in an offline state. Consequently, during the aggregation process on the server side, \(Straggler_2\) can dampen the variation of the intermediary model \( p_i^k \) and may even steer the shared intermediary model in the wrong direction, which in turn negatively impacts the parameter updates of the specified \( client_i \). Therefore, \(Straggler_2\)  will upload a lower tFIM to the server.

\subsubsection{Main Algorithm of FedSAF}

The steps of the FedSAF framework are formalized in Algorithm~\ref{alg:FedSAF}, which describes the detailed processes of model splitting, communication rounds, and parameter aggregation.

\begin{algorithm}
\caption{FedSAF}
\label{alg:FedSAF}
\begin{algorithmic}[1]

\STATE \textbf{Step 1: Model Splitting}
    \STATE \textbf{Input:} $m$ (number of clients), $\tau_i$ (complete model parameters of $client_i$), $\eta_i$ (personalized head layers of $client_i$)
    \STATE $lr$ (learning rate), $K$ (communication rounds)
    \STATE $Y_i$ (true labels for the training data of $client_i$) 
    \STATE \textbf{Calculate shared layers:} $\rho_i = \tau_i - \eta_i$ using Equation~\hyperref[eq:base_calculation_appendix]{(\ref*{eq:base_calculation_appendix})} 

    \STATE \textbf{Output:} $\eta$ (personalized model parameters), $\rho$ (updated shared layers)
    \STATE $\eta_i$ (updated personalized head layers), $p_i$ (a linear combination of model parameter sets adjusted by AMP and FIM)

\STATE \textbf{Step 2: Communication Rounds}
\FOR{each communication round $k = 1, 2, \ldots, K$}
    \STATE \textbf{Server-side updates:}
    
    \STATE Collect the shared layers $\rho^{k-1}_1, \ldots, \rho^{k-1}_m$ and the list of tFIM from the clients.
    
    AMP: Compute similarity coefficient $\xi_{i,j}$ based on equation ~\hyperref[eq:xi_Euclidean_base]{(\ref*{eq:xi_Euclidean_base})} and aggregate $z^k_i $ based on equation ~\hyperref[eq:mki_calculation]{(\ref*{eq:mki_calculation})}.
    
    \STATE 
    FIM: Compute weighted average of tFIM $\gamma_i$ based on Equation~\hyperref[eq:tfim_weight]{(\ref*{eq:tfim_weight})} and aggregate $p_i^k$ based on Equation~\hyperref[eq:pk_calculation]{(\ref*{eq:pk_calculation})}.

    \STATE The server sends $p^k_1, \ldots,p^k_m $ back to clients.
    
    \STATE \textbf{Client-side updates:}
    \STATE Each client $C_i$ updates its personalized head $\eta_i$ using Equation~\hyperref[eq:head_update]{(\ref*{eq:head_update})}.
    
    \STATE Then, update shared layers $\rho$ using the adjusted weights $p_i^k$ according to Equation~\hyperref[eq:total_layers_update]{(\ref*{eq:base_layers_update})}.

    \STATE Finally, update the list of tFIM based on equation ~\hyperref[eq:fim_trace]{(\ref*{eq:fim_trace})}.

\ENDFOR

\end{algorithmic}
\end{algorithm}


\paragraph{FedSAF-Server}

As in Algorithm ~\ref{alg:FedSAF}, in the server, there are two steps of aggregation of global models. The first aggregation is based on AMP. During the first aggregation, $z_i^k$(shown as Equation ~\ref{eq:mki_calculation}), a mediating variable in the global model, is calculated by the similarity between different clients. Then the second aggregation occurs and another mediating variable $p_i^k$(shown as Equation ~\ref{eq:pk_calculation} ) is created by the method FIM. Now, all equations related to these two-step aggregation are listed as follows:

\textbf{Step1: Aggregated by AMP}

$\xi_{i,j}$ is the similarity measure between client $i$ and client $j$, using model parameters from the previous round $k-1$ (shown as Equation ~\ref{eq:xi_manhattan_base}). It denotes the similarity measure and hence the weight assigned to the model
parameters from client j when updating client $i$'s model.

And then, the coef of self-client$_{i}$ can be calculated by:
\begin{equation}
    \xi_{ii} = 1 - \sum^{m}_{j \neq i} \xi_{ji}
    \label{eq:xi_relation_coef}
\end{equation}

 For the equation ~\ref{eq:xi_relation_coef}, $\sum^{m}_{j \neq i} \xi_{ji}$ is the sum of weights assigned to the model parameters of all other clients except client $i$, used to scale down the weight of client $i$'s own parameters to ensure that the sum of all weights equals 1.

 With the coef of self-client$_{i}$ and the coef of other client$_{j}$ $\xi_{i,j}$, the aggregation with AMP can be calculated and the global parameters can be expressed as follows:

If not splitting model:
\begin{equation}
    z_i^k = \xi_{ii} \tau^{k-1}_i + \sum^{m}_{j \neq i} \xi_{ji} \tau^{k-1}_j
    \label{eq:mki_calculation_total}
\end{equation}

If splitting model:
\begin{equation}
    z_i^k = \xi_{ii} \rho^{k-1}_i + \sum^{m}_{j \neq i} \xi_{ji} \rho^{k-1}_j
    \label{eq:mki_calculation}
\end{equation}


\textbf{Step2: Aggregated by FIM}

 $\gamma_i$ is the updated weight based on the trace of the Fisher Information Matrix for client $i$, which is adjusted by subtracting the sum of the weights of all other clients. It can be calculated by the equation ~\ref{eq:c}:

\begin{equation}
\begin{split}
    \gamma_i = 1 - \sum_{j \neq i}^{m} \gamma_j 
\end{split}
\label{eq:c}
\end{equation}


\begin{equation}
    p_i^k = \gamma_i \times z_i^k+ \sum^{m}_{j \neq i}\gamma_j \times z_j^k
    \label{eq:pk_calculation}
\end{equation}


The whole two-step aggregation equations are shown as follows: 

If not splitting model:
\begin{equation}
\begin{split}
    p_i^k =  \gamma_i \left( \xi_{ii} \tau^{k-1}_i + \sum_{j \neq i}^{m} \xi_{ji} \tau^{k-1}_j \right) + 
    \sum_{j \neq i}^{m} \gamma_j \left( \xi_{jj} \tau^{k-1}_j + \sum_{l \neq j}^{m} \xi_{lj} \tau^{k-1}_l \right)
\end{split}
\label{eq:pk_updated_full}
\end{equation}

If splitting model:
\begin{equation}
\begin{split}
    p_i^k =  \gamma_i \left( \xi_{ii}  \rho^{k-1}_i + \sum_{j \neq i}^{m} \xi_{ji}  \rho^{k-1}_j \right) + 
    \sum_{j \neq i}^{m} \gamma_j \left( \xi_{jj}  \rho^{k-1}_j + \sum_{l \neq j}^{m} \xi_{lj}  \rho^{k-1}_l \right)
\end{split}
\label{eq:pk_updated_base}
\end{equation}


After that, the server sends new global parameters $p_i^k$ back to the client$_{i}$. 

\paragraph{FedSAF-Client}
As in Algorithm~\ref{alg:FedSAF}, in the client, the parameters will be updated based on the regularization, which comprehensively considers globalization and personalization, and uses the Euclidean distance formula.

\textbf{One-step updating}

If the model splitting function is turned off, all parameters of the local model will be updated together, and the formula is as follows:

\begin{equation}
    \tau^{k+1}_i = \arg \min \left( F_i(\tau_i) + \frac{\lambda}{2\alpha_K} \|\tau_i - p_i^k\|^2 \right)
    \label{eq:total_layers_update} 
\end{equation}


\textbf{Two-step updating}

If using the function of model splitting, the head layers $\eta_i^{k}$, layers that do not participate in sharing, will be updated first, and then the base layers $\rho^{k+1}_i$, that are shared layers,  will be updated under the influence of regularization.

\textbf{step 1:} Freeze the base layer parameters and only update the head with SGD
\begin{equation}
    \eta_i^{k} = \eta_i^{k-1} - lr \cdot \nabla f_i(\eta_i^{k-1};  \rho^{k}_i)
    \label{eq:head_update}
\end{equation}


\textbf{step 2:}Freeze the head layer parameters and only update the base layer parameters

\begin{equation}
    \rho^{k+1}_i = \arg \min \left( F_i(\rho_i) + \frac{\lambda}{2\alpha_K} \|\rho_i - p_i^k\|^2 \right)
    \label{eq:base_layers_update} 
\end{equation}


After that, $\rho^{k+1}_i$, as the parameter of the new round of shared layer, will participate in the aggregation of the global model of the server in the $k+1$ round.

\section{Experiments}

\subsection{Data Collection}
In this paper, two primary gastric cancer datasets were utilized, SEED and BOT, to develop a federated learning model aimed at addressing gastric cancer recognition. To assess the model's generalization capabilities, two widely recognized datasets, FashionMNIST and CIFAR-10, were also employed. These datasets differ in key characteristics:

\begin{itemize}
  \item \textbf{Gastric Cancer Datasets (SEED and BOT):} Focused on detecting potential cancerous regions in histopathology images, these datasets are binary, containing two classes—normal and abnormal—reflecting real-world diagnostic scenarios.
  \item \textbf{FashionMNIST and CIFAR-10:} Designed to evaluate classification performance, these datasets encompass a larger number of classes, thereby testing the model's ability to distinguish between multiple categories.
\end{itemize}

Details of these datasets are shown in Table~\ref{tab:original_dataset_comparison}.
\begin{table}[!ht]
    \centering
    \caption{Original Datasets Analysis}
    \begin{tabular}{lcccc}
        \toprule
        \textbf{Features} & \textbf{SEED} & \textbf{BOT} & \textbf{FashionMNIST} & \textbf{CIFAR-10} \\
        \midrule
        Number of Classes       & 2          & 2           & 10           & 10 \\
        Sample Size    & 1770       & 700         & 70000        & 60000 \\
        Number of Clients       & 5          & 10          & 20           & 20 \\
        Image Size (Pixels)              & 512*512    & 2048*2048   & 28*28        & 32*32 \\
        Color Channel           & 3          & 3           & 1            & 3 \\
        Data Format             & PNG        & TIFF, SVG   & uint8        & uint8 \\
        \bottomrule
    \end{tabular}
    \label{tab:original_dataset_comparison}
\end{table}


Patient privacy concerns limit medical sample availability, causing varying sample sizes across institutions. Additionally, client willingness to participate in federated learning varies, leading to different numbers of cooperating entities. Disparities in imaging devices also result in variations in image size, color channels, and formats. To mirror these real-world conditions, these four datasets were selected to evaluate the effectiveness and practicality of the federated learning model.
 
\subsubsection{SEED and BOT Datasets} 

Upon reviewing professional gastric cancer databases, two datasets containing original histopathology images were identified. The first dataset, known as SEED, originates from the Second Jiangsu Big Data Development and Application Competition (Medical and Health Track) \cite{seed2020challenge}. This dataset comprises 1,770 histopathology images accompanied by a corresponding grayscale mask image that delineates cancerous regions.


Another dataset, referred to as BOT, comprises 700 histopathology images of gastric cancer obtained from the "Brain of Things: AI Challenge of Pathological Section Identification" \cite{pathological2017challenge}. This dataset includes 560 images labeled as abnormal, along with their corresponding mask images, and 140 as normal. 

\subsubsection{FashionMNIST and CIFAR-10 Datasets}

To test the generalization ability of the federated learning model, two additional datasets commonly used for classification tasks were also prepared.
FashionMNIST is a dataset comprising 28×28 grayscale images of 70,000 fashion products across 10 categories, with 7,000 images per category.
And CIFAR-10 consists of 60,000 32×32 color images in 10 classes, with 6,000 images per class.


\subsection{Data Processing} 

\begin{figure}[htb]
    \centering
    \includegraphics[width=1\textwidth]{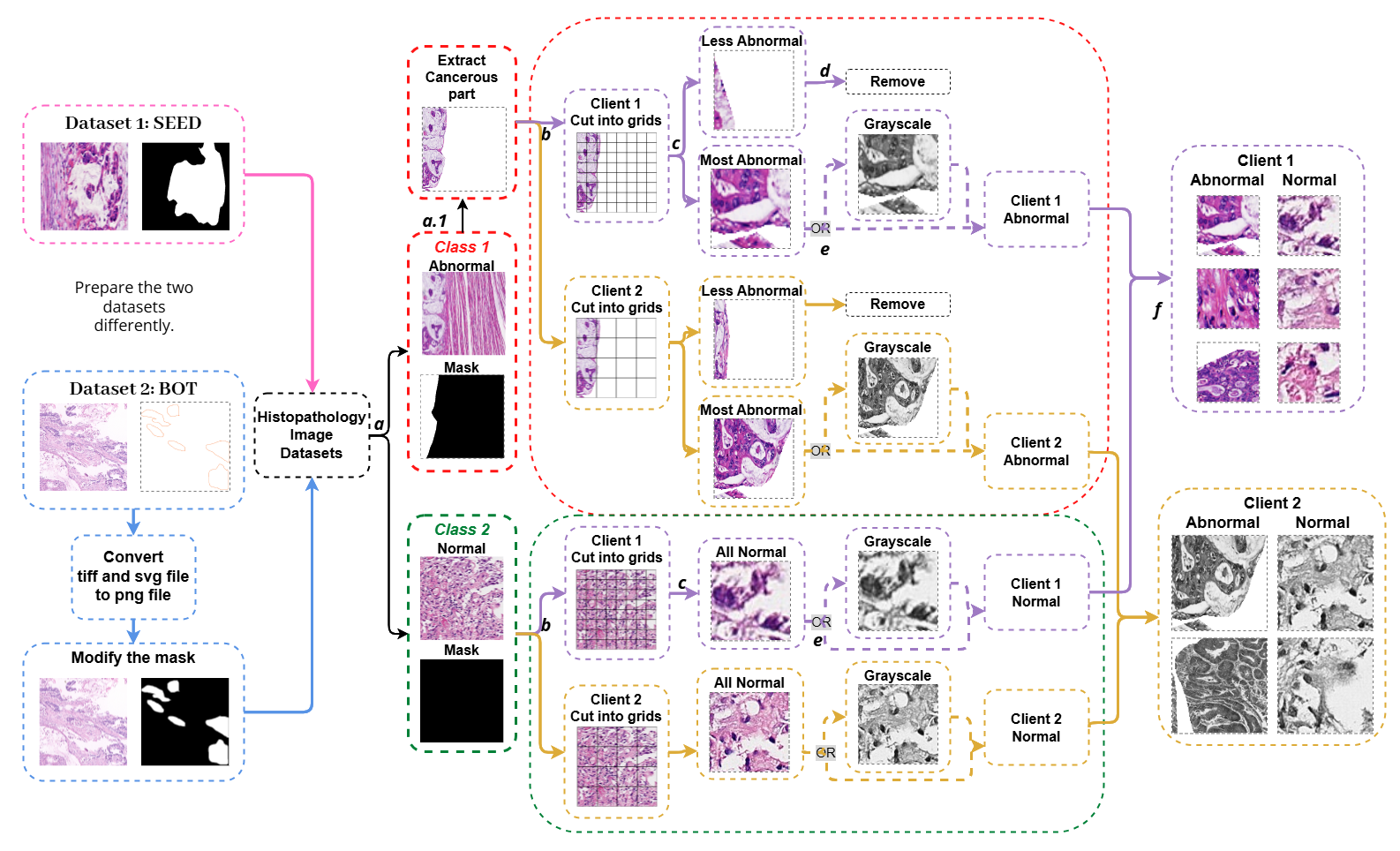}
    \caption{Data Processing}
    \label{fig:data_process}
\end{figure}

To enhance the realism of the federated learning model, 
developing a new data processing system is essential for constructing an effective and practical model, as illustrated in Figure~\ref{fig:data_process}.

Initially, all images were transformed into the same format, PNG, which effectively reduced the resources required for the following processing procedures.
Then, the images in the SEED dataset were classified based on their corresponding mask images. If all pixels in the mask image were black, the corresponding histopathology image was classified as normal; otherwise, it was categorized as abnormal.



Then, cancerous regions in the histopathology images would be extracted based on the mask images for the abnormal class in \textbf{Step a.1}. 
All abnormal areas would be transferred to the new image, while the rest of the area, would be white.

Next, the histopathology images would be randomly divided among different clients with varying sizes.
The following step, \textbf{Step b}, involved cropping the images to the specified size for both the abnormal and normal classes. 
After cropping, the abnormal images were analyzed based on the proportion of white area, which represented the removed normal tissue in the image. 
This process ensured that the images mostly contained abnormal tissue, helping the model effectively learn the features of the abnormal class and avoid misinterpreting normal patterns as abnormal.
To increase data heterogeneity, Some client-specific images might need to be converted to grayscale based on input parameters.

Finally, the datasets for multiple clients were prepared, with the flowchart depicting the case of two clients with differing sample sizes, image sizes, and color formats.





\subsection{ Experiments Setup } 

These experiments are based on the SEED dataset, which aims to verify the effectiveness and performance of different models and methods in the Federated Learning (FL) framework.

\subsubsection{DNN Selection}
To effectively train and optimize models for task-specific goals in the Federated Learning (FL) framework, it is essential to select an appropriate deep neural network (DNN) model to embed within the FedSAF framework. This paper evaluates four models: AlexNet, ResNet18, EfficientNet B0, and MobileNet V3 Small. These models were compared across various scenarios, including client-side local datasets (Separate), the FedAMP framework, and the FedSAF framework. The datasets were divided into IID (Independent and Identically Distributed) and non-IID (non-Independent and non-Identically Distributed) categories. The comparison involved six different cases.

\begin{table}[h!]
    \centering
    \caption{DNN Models Comparative Experiment}
    \label{DNN tabel}
    \resizebox{\textwidth}{!}{ 
    \begin{tabular}{lcccccc}
    \toprule
    & \multicolumn{2}{c}{\textbf{Separate}} & \multicolumn{2}{c}{\textbf{FedAMP}} & \multicolumn{2}{c}{\textbf{FedSAF}} \\
    \cmidrule(lr){2-3} \cmidrule(lr){4-5} \cmidrule(lr){6-7}
    & \textbf{IID} & \textbf{Non-IID} & \textbf{IID} & \textbf{Non-IID} & \textbf{IID} & \textbf{Non-IID} \\
    \midrule
    AlexNet & 0.5809 & 0.4661 & 0.7897 (104.4s) & 0.9279 (259.7s) & 0.9296 (98.2s) & 0.9307 (250.2s) \\
    ResNet 18 & 0.5339 & 0.4191 & 0.9948 (130.6s) & 0.9221 (303.6s) & 0.9968 (199.9s) & 0.9932 (454.8s) \\
    EfficientNet B0 & 0.4457 & 0.5286 & 0.9813 (168.7s) & 0.9245 (422.0s) & 0.9986 (292.8s) & 0.9944 (653.5s) \\
    MobileNet V3 Small & 0.5339 & 0.4191 & 0.9874 (104.3s) & 0.9583 (272.4s) & \textbf{0.9945 (155.8s)} & \textbf{0.9898 (356.0s)} \\
    \bottomrule
    \end{tabular}
    }
\end{table}

As shown in Table~\ref{DNN tabel}, the training performance of models within the FL framework is superior to that of local independent training. EfficientNet B0, ResNet18, and MobileNet V3 Small achieve accuracy greater than 0.9 on both IID and non-IID datasets. Although AlexNet embedded in the FL framework exhibits much lower training time compared to other models, whether on IID or non-IID datasets, it falls short in terms of accuracy compared to the other models.

MobileNet V3 Small not only achieves exceptional accuracy (up to 0.9945 on IID and 0.9898 on non-IID datasets in the FedSAF framework) but also maintains a competitive training time, making it the most balanced and efficient choice among the tested models. Therefore, MobileNet V3 Small will be used in subsequent experiments as the embedded DNN model in the FedSAF framework.

\subsubsection{Ablation Experiment}
To verify the effectiveness of the components used, an ablation experiment was conducted, comparing four scenarios: disabling both FIM and Model Splitting in FedSAF, disabling only Model Splitting, disabling only FIM, and using FedSAF with both components.

\begin{table}[h!]
    \centering
    \caption{Ablation Study Results}
    \label{table:Ablation Study Results}
    \resizebox{\textwidth}{!}{ 
    \begin{tabular}{cc|ccc|cc|c}
    \toprule
    \multicolumn{2}{c|}{\textbf{Components}} & \multicolumn{3}{c|}{\textbf{Averaged}} & \multicolumn{2}{c|}{\textbf{Std}}  & \multicolumn{1}{c}{\textbf{Parameters}} \\
    \textbf{FIM} & \textbf{Model Splitting} & \textbf{Training Loss} & \textbf{Test ACC} & \textbf{Test AUC} & \textbf{Test ACC} & \textbf{Test AUC} & \textbf{Transmitted} \\
    \midrule
     & & 0.1756 & 0.9583 & 0.9885 & 0.1362 & 0.095 & 2118690 \\
    $\checkmark$ &  & 0.0526 & 0.9799 & 0.9934 & 0.1252 & 0.0936 & 2118690 \\
     & $\checkmark$ & 0.0639 & 0.9709 & 0.9901 & 0.0851 & 0.0439 & 345400 \\
    $\checkmark$ & $\checkmark$ & 0.0558 & 0.9756 & 0.9914 & 0.1031 & 0.0787 & 345400 \\
    \bottomrule
    \end{tabular}
    }
\end{table}

Table~\ref{table:Ablation Study Results} shows that disabling Model Splitting improves Averaged Test Accuracy, while disabling FIM significantly reduces the number of parameters transmitted. The use of both FIM and Model Splitting components in FedSAF results in improved accuracy and reduced transmission.

Further analysis was conducted with line graphs comparing  Averaged Test Accuracy, and Std Test Accuracy for each combination of components across rounds.



\begin{figure}[ht]
    \centering
    \includegraphics[width=0.9\textwidth]{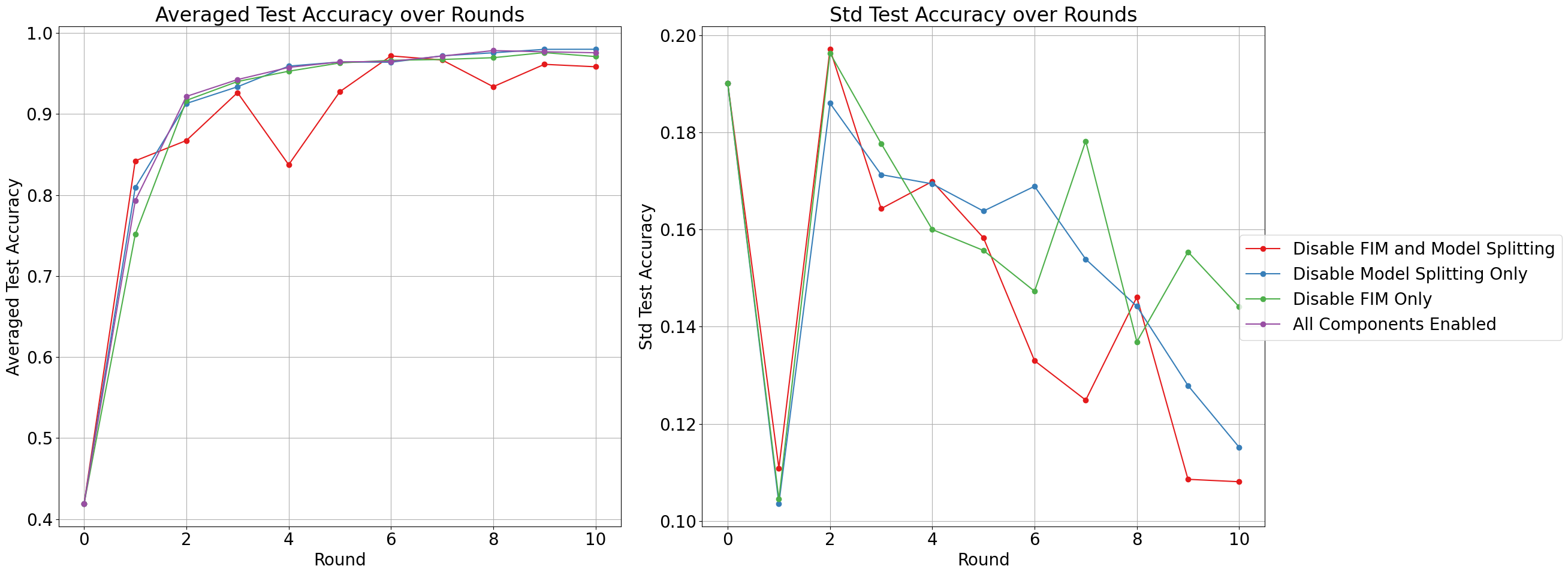}
    \caption{Averaged and Std Test Accuracy over Rounds for Different Component Combination}
    \label{fig:Discussion Ablation Averaged and Std Test Accuracy}
\end{figure}

The visualization of Averaged Test Accuracy in Figure~\ref{fig:Discussion Ablation Averaged and Std Test Accuracy} shows that while FedSAF with both FIM and Model Splitting disabled can achieve an accuracy of over 0.9, it remains unstable. The inclusion of the Model Splitting component helps mitigate this instability, suggesting that it enhances the stability of the global model on each client by preserving personalized layers. Comparing the results of FedSAF with only FIM disabled and the full FedSAF framework reveals that incorporating the Model Splitting component does not degrade the model's performance.


In addition, the visualization of Standard Deviation Test Accuracy in Figure~\ref{fig:Discussion Ablation Averaged and Std Test Accuracy} compares the results when disabling both the FIM and Model Splitting components with the full FedSAF framework. It demonstrates that using FedSAF with both FIM and Model Splitting reduces the standard deviation of Test Accuracy, while the model without these components exhibits larger fluctuations. This indicates that these two components effectively enhance the robustness of the FL framework.

In summary, the combination of the FIM Component and Model Splitting Component in FedSAF effectively mitigates the accuracy fluctuations caused by data distribution differences between clients. Therefore, these added components contribute positively to the model's performance.

\subsubsection{Hyperparameter Tuning}  

The Parameter Tuning experiment focuses on enhancing the performance of the FedSAF framework when processing non-IID data by selecting the appropriate Distance type and adjusting the number of Head layers to be frozen.


The hyperparameter tuning for Distance Types in the Attention Message Passing (AMP) component helps select the optimal distance metric for measuring client similarities. As shown in Table~\ref{table:Distance Metrics Performance}, among the three distance methods evaluated, Manhattan distance achieves the lowest training loss and high test accuracy with minimal fluctuations, indicating that it provides the most stable and robust performance. Moreover, the Averaged Train Loss for Cosine distance is noticeably higher than that of both Manhattan and Euclidean distances. This indicates its ineffectiveness in measuring similarity between clients, leading to poor convergence.

    \begin{table}[ht!]
        \centering
        \caption{Distance Metrics Performance}
        \label{table:Distance Metrics Performance}
        \begin{tabular}{lccc}
        \toprule
        \textbf{Distances} & \textbf{Averaged Train Loss} & \textbf{Averaged Test Accuracy} & \textbf{Std Test Accuracy} \\
        \midrule
        Cosine & 3.6636 & 0.9756 & 0.1081 \\
        Euclidean & 0.0459 & 0.9798 & 0.1152 \\
        \midrule
        \textbf{Manhattan} & \textbf{0.0446} & \textbf{0.9786} & \textbf{0.1441} \\
        \bottomrule
        \end{tabular}
    \end{table}

\begin{figure}[ht]
        \centering
        \includegraphics[width=0.9\textwidth]{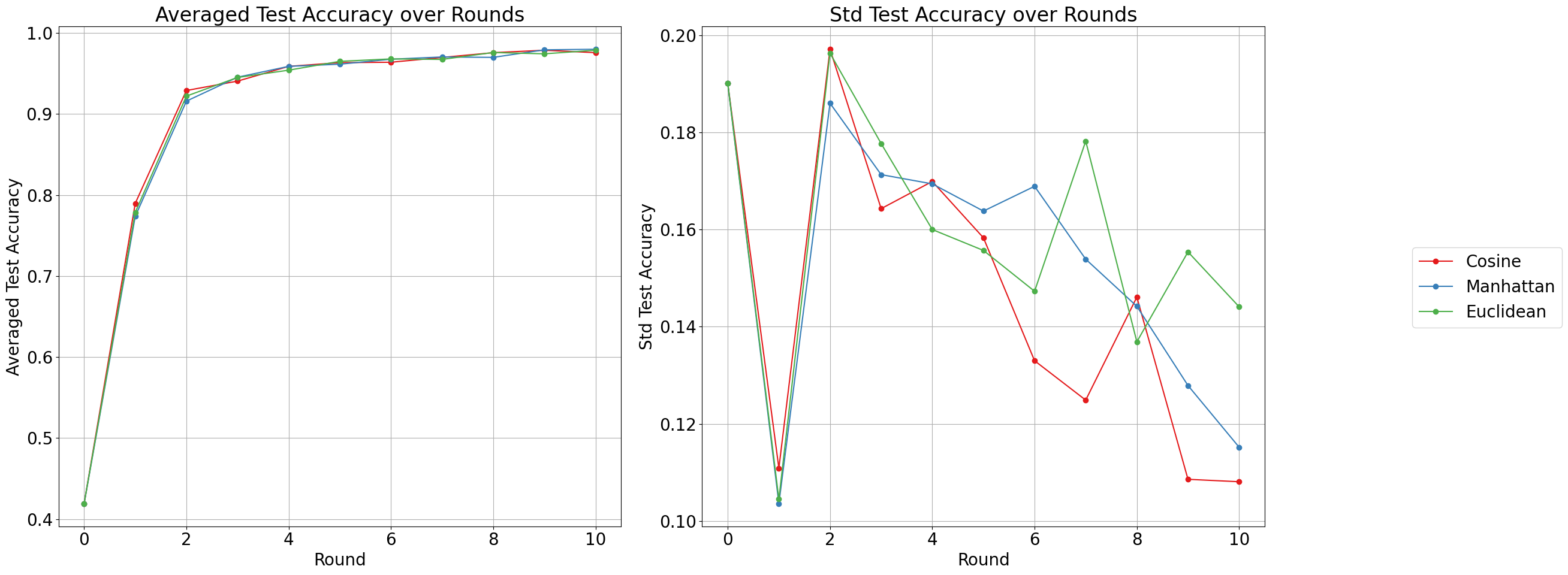}
        \caption{Averaged and Std Test Accuracy over Rounds for Different Distances}
        \label{fig:Discussion Distance Averaged and Std Test Accuracy}
    \end{figure}

Further comparison of Averaged Test Accuracy and Standard Deviation Test Accuracy reveals that while both Manhattan and Euclidean distances exhibit stable performance, Manhattan distance demonstrates superior stability with less fluctuation in the standard deviation, indicating its better robustness (see Figure \ref{fig:Discussion Distance Averaged and Std Test Accuracy}).
Based on these findings, Manhattan distance was selected for the final parameter tuning in the AMP component due to its overall performance, stability, and robustness.


In tuning the Number of Head Layers in the Model Splitting component, the paper evaluates five different configurations, ranging from 0 to 12 frozen layers. As shown in Table~\ref{tabel:Hyperparameter Tuning for Head Layers}, increasing the number of Head Layers results in a gradual decrease in transmitted parameters, but the model accuracy remains high across all configurations. Notably, when 6 layers are frozen, the total number of transmitted parameters drops significantly to 345,400—less than the total transmitted by FedAvg, while still maintaining good accuracy.

\begin{figure}[ht]
        \centering
        \includegraphics[width=0.9\textwidth]{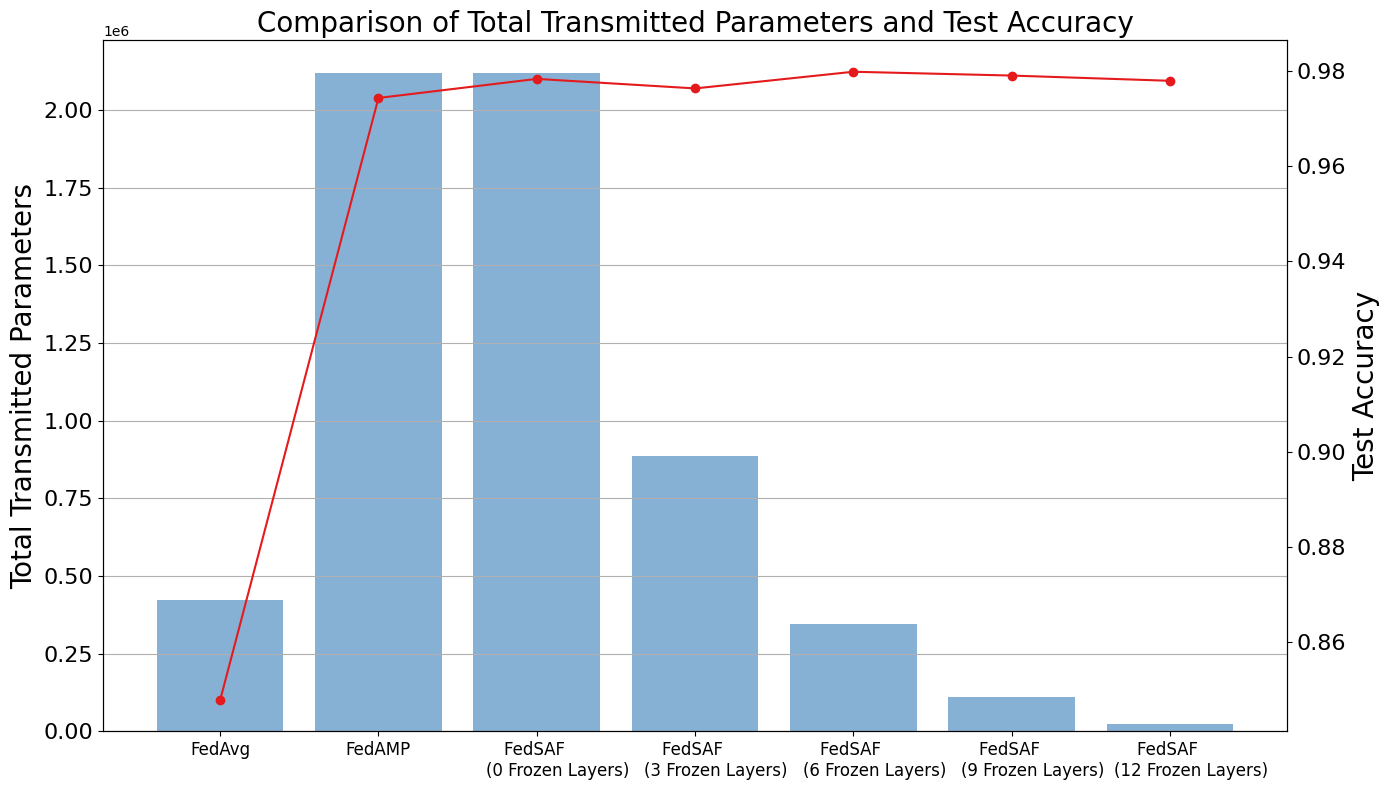}
        \caption{Comparison of Total Transmitted Parameters and Test Accuracy}
        \label{fig:Discussion Number of head}
    \end{figure}

Figure~\ref{fig:Discussion Number of head} highlights that when no layers are frozen, the transmitted parameters are approximately six times higher than when six layers are frozen.

Given the trade-off between reduced communication costs and preserved model performance, the final choice for the Model Splitting Component was to freeze 6 head layers.

    \begin{table}[ht!]
        \centering
        \caption{Hyperparameter Tuning for Head Layers}
        \label{tabel:Hyperparameter Tuning for Head Layers}
        \begin{tabular}{lccc}
        \toprule
        \textbf{Framework} & \textbf{Head layers} & \textbf{Total parameters transmitted} & \textbf{Averaged Test accuracy} \\
        \midrule
        FedAvg & 0 & 423738 & 0.8479 \\
        FedAMP & 0 & 2118690 & 0.9743 \\
        \midrule
        FedSAF & 0 & 2118690 & 0.9783 \\
        FedSAF & 3 & 887320 & 0.9763 \\
        \textbf{FedSAF} &\textbf{6} &\textbf{345400} & \textbf{0.9798} \\
        \vspace{-4.4 ex} \\ 
        \hdashrule[0.05ex]{2cm}{0.5pt}{2pt 2pt} & \hdashrule[0.05ex]{2cm}{0.5pt}{2pt 2pt} & \hdashrule[0.05ex]{2cm}{0.5pt}{2pt 2pt} & \hdashrule[0.05ex]{2cm}{0.5pt}{2pt 2pt} \\ 
        FedSAF & 9 & 111320 & 0.9790 \\
        FedSAF & 12 & 24160 & 0.9792 \\
        \bottomrule
        \end{tabular}
    \end{table}

\subsection{Results}
Table \ref{noniidgeneralization} presents the results of generalization experiments conducted on three additional datasets—BOT, FashionMNIST, and CIFAR-10—to further evaluate the performance of the FedSAF model in non-IID environments. These experiments aimed to assess the model's ability to generalize across various data distributions and task complexities. 

The FedSAF model was compared with several other federated learning approaches to evaluate their generalization performance. The results, shown in Table \ref{noniidgeneralization}, highlight the highest accuracy achieved by FedAvg, FedProx, FedREP, FedAMP, and FedSAF across the datasets. FedSAF demonstrated strong performance with high accuracy on most datasets, showcasing its robustness and adaptability to heterogeneous and unbalanced data distributions. However, on certain datasets, particularly CIFAR-10, FedAMP slightly outperformed FedSAF, suggesting that model performance may vary depending on the dataset characteristics.

Overall, FedSAF exhibits superior generalisation capabilities and highly consistent performance, enabling it to achieve consistent training results in a federated learning environment. This superior performance makes FedSAF particularly suitable for dealing with non-independently and identically distributed data distributions, and its stable optimisation path and reduced standard deviation fluctuations allow for consistent performance across clients, thus ensuring cross-client consistency and robustness. This makes FedSAF a reliable choice for performing federated learning in heterogeneous data environments and maintains high performance even in the face of the diversity and unevenness of data distributions.

\begin{table}[!ht]
\centering
\caption{Noniid-Unbalanced Dataset Generalization: Best\_Accuracy}
\label{noniidgeneralization}
\begin{tabular}{lcccc}
\toprule
\textbf{Method} & \textbf{SEED} & \textbf{BOT} & \textbf{FashionMNIST} & \textbf{Cifar10} \\
(clients n, classes per client S) & (5, 2) & (10, 2) & (20, 10) & (20, 10) \\
\midrule
FedAvg \cite{mcmahan2017communication}& 0.7362 & 0.5871 & 0.6677 & 0.1741 \\
FedProx \cite{li2020federated}& 0.7953 & 0.6672 & 0.6654 & 0.2036 \\
\cmidrule(lr{0.15pt}){1-5}
FedRep \cite{husnoo2023fedrep} & 0.8937 & 0.6202 & 0.9586 & 0.8377 \\
FedAMP \cite{huang2021personalized}& 0.9790 & 0.7834 & 0.9622 & 0.8400 \\
\cmidrule(lr{0.15pt}){1-5}
\textbf{FedSAF (ours)} & \textbf{0.9843} & \textbf{0.8116} & \textbf{0.9642} & \textbf{0.8380} \\
\bottomrule
\end{tabular}
\end{table}

\subsection{Discussion}
The analysis above demonstrates that FedSAF excels in generalization and consistency across multiple datasets, particularly showcasing its robustness in handling non-independently and identically distributed (non-IID) data. FedSAF’s superior performance enabled it to achieve the highest accuracy on the SEED, BOT, and FashionMNIST datasets, highlighting its adaptability in managing heterogeneous and unbalanced data distributions. However, on the CIFAR-10 dataset, although FedSAF's performance remains near optimal, FedAMP outperforms it slightly.

This performance disparity can be attributed to the characteristics of the CIFAR-10 dataset. The images in this dataset are more complex and diverse, placing higher demands on the model's learning and generalization capabilities. In such cases, FedAMP may have adapted better to the heterogeneity of the dataset through a personalized aggregation strategy after initial fluctuations, whereas FedSAF maintained relatively stable performance. This suggests that different federated learning approaches may excel on specific datasets, depending on how well the model can address dataset heterogeneity and complexity.

Given the typically small sample sizes in gastric cancer datasets, data enhancement techniques have been incorporated in this project. Both grayscale and color gastric cancer images can be identified and cropped to help expand the sample size for each client’s local dataset. This approach not only increases the sample size but also ensures the protection of client data privacy, leading to improved training outcomes.

Data and system heterogeneity are common challenges in federated learning. To address these issues, Attention Message Passing (AMP) and Fisher Information Matrix (FIM) methods have been introduced. AMP allows the model to dynamically adjust the weight of each client based on their contribution, while FIM identifies and processes stragglers. These techniques help alleviate model offset and correct stragglers, ultimately improving the accuracy of model predictions.

Each weighted aggregation in the server typically causes the client model to drift towards the global model. To mitigate this, the model splitting method is employed, which allows clients to perform local stratification and freeze the weights of certain layers without uploading them to the central server. This enables clients to retain their unique data features, allowing for a balance between global and personalized models. As a result, this method helps improve the performance degradation that can occur due to excessive globalization.

High communication costs arise from the frequent uploading of numerous model parameters. To reduce these costs, the model splitting technique is also introduced, limiting the upload to only a subset of parameters rather than the entire model. This method helps to significantly lower communication costs, making it more efficient in federated learning environments.

\section{Conclusion}

This paper introduces FedSAF, a novel federated learning framework designed to enhance gastric cancer detection accuracy while maintaining data privacy across medical institutions. FedSAF addresses challenges associated with small sample sizes and data heterogeneity through personalized federated learning. The framework incorporates several innovative components, including Model Splitting, Attentive Message Passing (AMP), and the Fisher Information Matrix (FIM), to reduce transmission costs, enhance collaboration efficiency, and minimize inconsistencies among clients.

The main contributions are as follows:

\begin{itemize}
    \item \textbf{Model splitting mechanism}: Divide the model into fixed and personalized parts, improve data privacy, reduce communication costs, and adapt to data heterogeneity by only transmitting shallow parameters while retaining deep layers locally.
    \item \textbf{Attention message passing (AMP)}: Aggregate the similarity of models based on multiple distance metrics (such as Euclidean, Manhattan, and cosine distance) to improve client collaboration capabilities.
    \item \textbf{Fisher information matrix (FIM)}: Dynamically adjust the aggregation weight to improve the test accuracy from 95.83\% to 97.99\% and reduce the accuracy difference between clients.
\end{itemize}

Experiments on GasHisSDB, SEED, BOT, FashionMNIST, and CIFAR-10 datasets show that FedSAF can effectively handle heterogeneous and unbalanced data distributions. The test accuracy on the SEED dataset reached 98.43\%, exceeding FedAMP's 97.90\%; the test accuracy on the BOT dataset was 81.16\%, higher than FedAMP's 78.34\%.

Despite its excellent performance, FedSAF still has room for improvement: its unified model architecture assumption limits its adaptability to heterogeneous environments; synchronous updates may affect efficiency; and high personalization may lead to overfitting on IID data. Future research can explore heterogeneous model support, asynchronous update mechanisms, and adaptive personalization levels to achieve a better balance between generalization and personalization.

\bibliographystyle{unsrt} 
\bibliography{main}

\end{document}